\pdfoutput=1

\documentclass[11pt]{article}

\usepackage{EACL2023}

\usepackage{times}
\usepackage{latexsym}
\usepackage{enumerate}
\usepackage{enumitem}
\usepackage{makecell}
\usepackage{multirow}
\usepackage{booktabs}
\usepackage{soul}
\usepackage{amsmath}
\usepackage[T1]{fontenc}

\usepackage[utf8]{inputenc}

\usepackage{microtype}

\usepackage{inconsolata}

\usepackage{tikz-dependency}
\usepackage{graphicx}
\usepackage{verbatim}
\usepackage{tcolorbox}
\usepackage{xcolor}
\tcbuselibrary{skins}
\usetikzlibrary{shadings}
\usepackage{listings}
\definecolor{backcolour}{rgb}{0.965,0.965,0.965}
\usepackage[T1]{fontenc}
\usepackage[scaled=0.85]{beramono}            
\lstdefinestyle{mystyle}{
    backgroundcolor=\color{backcolour},
    basicstyle=\ttfamily\footnotesize\bfseries,
    breakatwhitespace=false,
    breaklines=true,
    captionpos=t,
    keepspaces=true,
    numbersep=5pt,
    showspaces=false,
    showstringspaces=false,
    showtabs=false,
    tabsize=2,
    frame=single,
    framerule=0.0mm,
    framesep=5pt,
    xleftmargin=5pt,
    xrightmargin=5pt,
}

\definecolor{mycolor}{RGB}{0, 102, 204}
\tcbset{enhanced,colback=red!5!white,
boxrule=0.1pt,
colframe=red!75!black,fonttitle=\bfseries}

\newtcolorbox{gptbox}{
  enhanced,               
  colback=mycolor!5!white, 
  colframe=mycolor,        
  fonttitle=\bfseries\color{white}, 
  title style={
    overlay,               
    outer arc=0mm,         
    colback=mycolor,       
    coltext=white,         
    halign=center,         
    valign=center,         
    before upper={\rule[2pt]{0.25cm}{0.5ex}\hspace{0.15cm}}, 
    after upper={\hspace{0.15cm}\rule[2pt]{0.25cm}{0.5ex}}, 
  },
  title={GPT Prompt 1},      
  sharp corners,           
  breakable,               
}

\definecolor{mygreen}{RGB}{0, 128, 0} 
\definecolor{fontcolor}{RGB}{255, 255, 255} 

\title{X-AMR Annotation Tool}

\author{Shafiuddin Rehan Ahmed \hspace{3mm} Jon Z. Cai \hspace{3mm} Martha Palmer \hspace{3mm} James H. Martin\\[0.5mm] University of Colorado, Boulder, USA \\[1mm]
\texttt{\{shah7567, jon.z.cai\}@colorado.edu}}

\begin{document}
\maketitle
\begin{abstract}
This paper presents a novel \textbf{Cross}-document \textbf{A}bstract \textbf{M}eaning \textbf{R}epresentation (X-AMR) annotation tool designed for annotating key corpus-level event semantics. Leveraging machine assistance through the Prodigy Annotation Tool, we enhance the user experience, ensuring ease and efficiency in the annotation process. Through empirical analyses, we demonstrate the effectiveness of our tool in augmenting an existing event corpus, highlighting its advantages when integrated with GPT-4. Code and annotations: \href{https://github.com/ahmeshaf/gpt_coref}{github.com/ahmeshaf/gpt\_coref}\footnote{Demo: \href{https://youtu.be/TuirftxciNE}{https://youtu.be/TuirftxciNE}} \footnote{Live Link: \href{http://eacldemo.acl-lawpaper34-demo.site/}{eacldemo.acl-lawpaper34-demo.site/}}
\end{abstract}

\newcommand{\argzero}{\text{ARG-0}}
\newcommand{\argone}{\text{ARG-1}}
\newcommand{\argL}{\text{ARG-Loc}}
\newcommand{\argT}{\text{ARG-Time}}

\section{Introduction}
Semantic representations of events play a pivotal role in natural language processing (NLP) tasks, facilitating the understanding and extraction of meaningful information from text. Among the various approaches to represent events, Semantic Role Labeling (SRL; \citet{palmer-etal-2005-proposition}) and Abstract Meaning Representation (AMR; \citet{banarescu-etal-2013-abstract}) have gained significant attention. In this paper, we delve into the realm of semantic event representations, with a particular focus on a method for expanding AMR.

AMR, a graph-based semantic representation, aims to capture the underlying meaning of sentences by breaking them down into atomic concepts and their semantic relationships. Each concept in AMR is associated with a unique identifier, and the relationships between concepts are represented as labeled edges in a graph. AMR has proven to be versatile, serving as a valuable resource for a wide range of NLP tasks such as machine translation, question answering \cite{fu-etal-2021-decomposing-complex}, and summarization \cite{liao-etal-2018-abstract}. Its ability to provide a structured, language-independent representation of textual content makes it an essential tool in the NLP toolkit.

However, despite its many merits, current AMR techniques are not without limitations. One of the primary challenges lies in linking temporal relations and entity coreference across sentences and documents. This limitation hinders the comprehensive understanding of text, as it often fails to capture the intricate interplay between events and entities that span multiple contexts. This issue becomes particularly pronounced in scenarios involving cross-document event coreference, where events mentioned in one document need to be linked to events in other documents for a coherent understanding of a larger narrative.

To illustrate the challenge of coreference across documents, consider the following example: Two news articles discuss a corporate acquisition. In one article, the event is described as "Company A's purchase of Company B on July 1st, 2008" while in another article, it is referred to as "In 7/08 Company B was acquired by Company A." Establishing the coreference relationship between these two descriptions is non-trivial, yet crucial for creating a comprehensive representation of the acquisition event.


\newcommand{\ecb}{\text{ECB+}}

To specifically address the intricate challenges of cross-document event coreference resolution, our research introduces two significant contributions. Firstly, we propose a novel framework X-AMR. This framework is an enhancement of the existing AMR, specifically designed to overcome the challenges inherent in linking events and entities across different documents. X-AMR effectively combines the strengths of AMR with the ability to create a more comprehensive and coherent depiction of narratives that span multiple sources.

Secondly, the development of a specialized interface is another key contribution of our work. Utilizing the model-in-the-loop annotation methodology, we have leveraged the customized Prodigy annotation tool to augment an existing event coreference dataset, the Event Coref Bank plus (\ecb; \citet{cybulska-vossen-2014-using}). This development has facilitaed the annotation of X-AMR representations, focusing on the annotation interface and the enhanced X-AMR dataset. Additionaly, we present an evaluation showcasing the accuracy and efficiency of our approach. Our research endeavors to demonstrate the effectiveness of X-AMR in addressing the limitations of current sentence level AMR, especially in linking temporal relations and entity coreference across sentences and documents.

\section{Related Work}
AMR is a formalism meticulously crafted to capture the semantic nuances of natural language expressions with versatile and expressive power. In the field of Natural Language Processing (NLP), automatic AMR parsing transforms natural language inputs into formal AMR representations, which have demonstrated utility in a diverse array of downstream applications such as Summarization \cite{liao-etal-2018-abstract}, Dialog systems \cite{ bonial-etal-2020-dialogue, bai-etal-2021-semantic}, Question-Answering \cite{kapanipathi-etal-2021-leveraging}, Machine Translation \cite{li-flanigan-2022-improving}, Language Modeling \cite{bai-etal-2022-graph},  and Fact Checking \cite{ribeiro-etal-2022-factgraph}.

Formally, AMR are structured as labeled, rooted, directed acyclic graphs, which capture abstract concepts, predicate-argument relationships, and entities found in sentences or utterances. They integrate the semantic content addressed by different representation schemes such as SRL, named entities recognition (NER; \citet{NNER}), and coreference resolution into a unified representation. For example, for sentence ``HP acquired EYPMCS.'', the corresponding AMR is:
\begin{lstlisting}
(d / acquire-01
    :ARG0 (c / company
        :name (n / name
            :op1 "HP"))
    :ARG1 (c2 / company
        :name (n2 / name
            :op1 "EYPMCS"))
\end{lstlisting}
The above AMR graph captures concepts such as events such as ``acquire'', named entities such as the HP company, and properties of the entity such as their names as graph nodes and subgraphs. Their interrelations between concepts and events are then depicted through labeled edges. Events are denoted using Propbank rolesets, and semantics relations of the entities and events are specified through numbered arguments and non-core relations from AMR's role inventory. For example, in the above acquisition event, the \texttt{ARG0} typically specifies the stereotypical agent of an event and \texttt{ARG1} typically specifies the stereotypical patient of an event. Additionaly, AMR graphs formalize local temporal information, as shown in the provided example.

In the preceding disucssion, we highlighted the expressiveness of AMR. However, the expressiveness of AMR introduces complexities in AMR annotation, historically a significant bottleneck for NLP community. The challenge has been to provide a substantial volume of AMR annotations to the data hunger statistical machine learning models given the limitations of available tools. The ISI editor, serving as the first AMR editor, has supported the AMR community for over a decade. Despite the efficacy of the ISI editor, its learning curve is notably steep for annotators. To make AMR annotation more accessible, \citet{jon-etal-2023-camra} developed a new annotation approach. They introduced an AMR editor based on coding, complemented by a neural network parser model, to streamline the annotation process.

The remarkable progress in large language model-based coding assistance, pioneered by OpenAI and Microsoft, is transforming the landscape of program synthesis in software engineering. These models, trained in both natural language and programming languages, excel at completing programs by intelligently integrating code history and human instructions. In a similar vein, CAMRA leverages these large language models (LLMs) to enhance AMR annotation. We are pioneering the extension of LLMs' capabilities, broadening their application to include more complex tasks such as cross-sentential and cross-document coreference and event linking. This initiative represents a significant step forward in harnessing the power of LLMs for even more sophisticated and long-distance dependent language processing tasks.

\begin{figure*}[t!]
    \centering
    \includegraphics[scale=0.75]{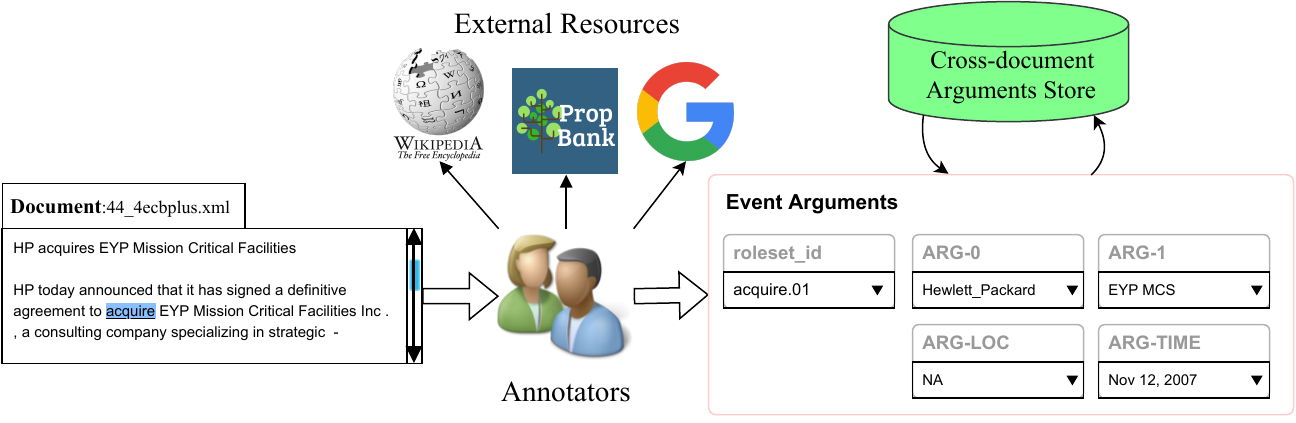}
    \caption{The Annotation Methodology of X-AMR. The annotators are presented with PropBank and are allowed to use external resources, such as Wikipedia and Google News, during the annotations.
    }
    \label{fig:anno-workflow}
\end{figure*}

\section{Annotation Methodology}
The annotation workflow, as depicted in Figure \ref{fig:anno-workflow}, comprises of two phases. In the first phase we annotate the roleset IDs of the event triggers. Then we specify the arguments of the event incrementally. During these two phases, we maintain an arguments store and a model-in-the-loop that queries the store and suggests annotators with the most likely arguments. This store and the model are updated when new events are annotated.

Next, we discuss the annotation guidelines, the interface, and the model-in-the-loop in the annotation workflow.
\subsection{Annotation Guidelines for X-AMR}
\vspace*{-1.5mm}
\label{sec:annotation}

We aim to annotate key event semantics with four arguments, \argzero, \argone, \argL, and \argT, capturing agent, patient (and theme), location, and temporal information. The selection of these arguments is to circumscribe an event by its \textit{minimal participants} \cite{lombard2019events, Guarino2022-GUAETN}.
We use the guidelines presented in the next section to hand annotate the roleset and argument information for the ECB+ train, development, and test sets using the standardized split of \citet{cybulska-vossen-2014-using}. Following the annotation guidelines, we provide the enriched annotations of the ECB+ corpus by two Linguistic students. We use a model-in-the-loop annotation methodology with the prodi.gy annotation tool.
\subsubsection{PropBank \& AMR}
Semantic role labeling (SRL) centers on the task of assigning the same semantic role to an argument across various syntactic constructions. For example, \textit{the window} can be the (prototypical) Patient, or thing broken, whether expressed as syntactic object (\textit{The storm broke the window}) or syntactic subject (\textit{The window broke in the storm}).


The Proposition Bank (PropBank; \citet{palmer-etal-2005-proposition,pradhan-etal-2022-propbank}) has over 11,000 Frame Files providing valency information (arguments and their descriptions) for fine-grained senses of English verbs, eventive nouns, and adjectives. Figure \ref{fig:enter-label} gives an example Frame File for \textit{agree} as well as an instantiated frame for \textit{HP has an agreement to acquire EYP}.



\vspace{1.5mm}


\begin{figure}
    \centering
\begin{minipage}[b]{0.46\columnwidth}
\noindent
\begin{verbatim}
agree.01 - agree
 ARG-0: Agreer
 ARG-1: Proposition

\end{verbatim}
\end{minipage}
\begin{minipage}[b]{0.01\columnwidth}
\hfill
\end{minipage}
\begin{minipage}[b]{0.43\columnwidth}
\begin{verbatim}
agree.01
 ARG-0: HP
 ARG-1: acquire.01
    ARG-1: EYP
\end{verbatim}
\end{minipage}

    \caption{The PropBank roleset definitions of agree.01 and the expected annotations in X-AMR.}
    \label{fig:enter-label}
\vspace*{-2mm}
\end{figure}

The resulting nested predicate-argument structures from PropBank style-SRL also form the backbones of AMRs, which in addition includes Named Entity (NE) tags and Wikipedia links (for `HP' and `EYP' in our example). AMRs also include explicit variables for each entity and event, consistent with Neo-Davidsonian event semantics, as well as inter- and intra-sentential coreference links to form directed, (largely) acyclic graphs that represent the meaning of an utterance or set of utterances.

Our enhanced X-AMR representation follows AMR closely with respect to NE and coreference, but stops short of AMR's additional structuring of noun phrase modifiers (especially with respect to dates, quantities and organizational relations), the discourse connectives and the partial treatment of negation and modality. However, we go further than AMR by allowing for cross-document coreference as well as multi-sentence coreference.  X-AMR thus provides us with a flexible and expressive event representation with much broader coverage than standard event annotation datasets such as ACE\footnote{https://www.ldc.upenn.edu/collaborations/past-projects/ace} or Maven \cite{wang-etal-2020-maven}.
\subsubsection{Roleset Sense Annotation}
The first step in the annotation process involves identifying the roleset sense for the target event trigger in the given text. Annotators, using an embedded PropBank website and the assistance of the tool's model, select the most appropriate sense by comparing senses across frame files.

\noindent \textbf{Handling Triggers with No Suitable Roleset:}
If there is no appropriate roleset that specifies the event trigger, particularly in cases when the trigger is a pronoun (it) or proper noun (e.g., Academy Awards), the annotator must then search for a roleset that defines the appropriate predicate.

\definecolor{hlcolor}{HTML}{DAE6FB}
\newcommand{\hlcyan}[1]{{\sethlcolor{hlcolor}\hl{#1}}}
\newtcolorbox{mybox3}[2][]{boxsep=1mm, arc=.15em, left=5pt,right=5pt,top=2pt,bottom=2pt,boxrule=0.1mm,
colbacktitle=black!45!white,colback=white, coltitle=white,
title={#2}, #1}
\definecolor{evtcolor}{HTML}{486485}
\newcommand{\highlightevt}[1]{\hlcyan{~\textbf{#1} {\textbf{\tiny {\color{evtcolor} EVT}}}~}}
\newcommand{\helvetext}[1]{\fontfamily{phv}\fontsize{7.5pt}{8pt}\selectfont  #1}
\newcommand{\helvetexttitle}[1]{\fontfamily{phv}\fontsize{8pt}{8pt}\selectfont \textbf{#1}}
\begin{figure}[t]
\begin{mybox3}{\helvetexttitle{Target Mention}}
\small
\tt
{HP today announced that it has signed a definitive \highlightevt{agreement} to acquire EYP Mission Critical Facilities Inc.}
\end{mybox3}
\includegraphics[scale=0.825]{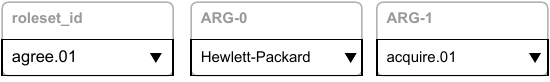}
\caption{Eventive ARG-1 in the roleset agree.01. The ARG-1 clause is annotated as the connecting event with roleset ID acquire.01}
\vspace*{-3.5mm}
\label{fig:eventive-arg}
\end{figure}

\subsubsection{Document-level Arguments Identification}
Next, we identify the document and corpus-level \argzero~and \argone~of the selected roleset. Annotators use the embedded PropBank website as a reference for the roleset's definition, ensuring that the \argzero~(usually the agent) and \argone~(typically the patient) are consistent with the roleset's constraints. For arguments that cannot be inferred, the annotators leave those fields empty.

\vspace{1.5mm}
\noindent \textbf{Within- and Cross-Document Entity Coreference Annotation:}
Annotators perform within- and cross-document entity coreference using a drop-down box of argument suggestions (suggested by the model-in-the-loop), simplifying coreference link establishment.

\vspace{1.5mm}
\noindent \textbf{Nested \argone:}
In many cases, the \argone~may itself be an event. In such cases, the annotator is tasked with identifying the head predicate of the \argone~role and providing its corresponding roleset ID. We then search for the annotations for such an \argone~and connect it to the target event. Fig \ref{fig:eventive-arg} has an example of a mention with an eventive \argone. For this, the annotator needs to provide the roleset for the  predicate of the \argone~clause (agree.01) as the \argone\space in this annotation process.
\vspace{-1mm}
\paragraph{\argL~\& \argT~Identification} Annotators may also utilize external resources, such as Wikipedia\footnote{Although we add this in the guidelines, the annotators do not wikify. Our choice is to use Wikipedia over the more commonly used KB-wikidata because of GPT-friendly identifiers of the pages. Check out Appendix \ref{sec:app-prompt}.}, or Google-News, for
the accurate identification of temporal and spatial arguments. This is required when the document does not explicitly mention the location and time of the event.

\begin{figure}[t]
    \centering
    \includegraphics[scale=0.75]{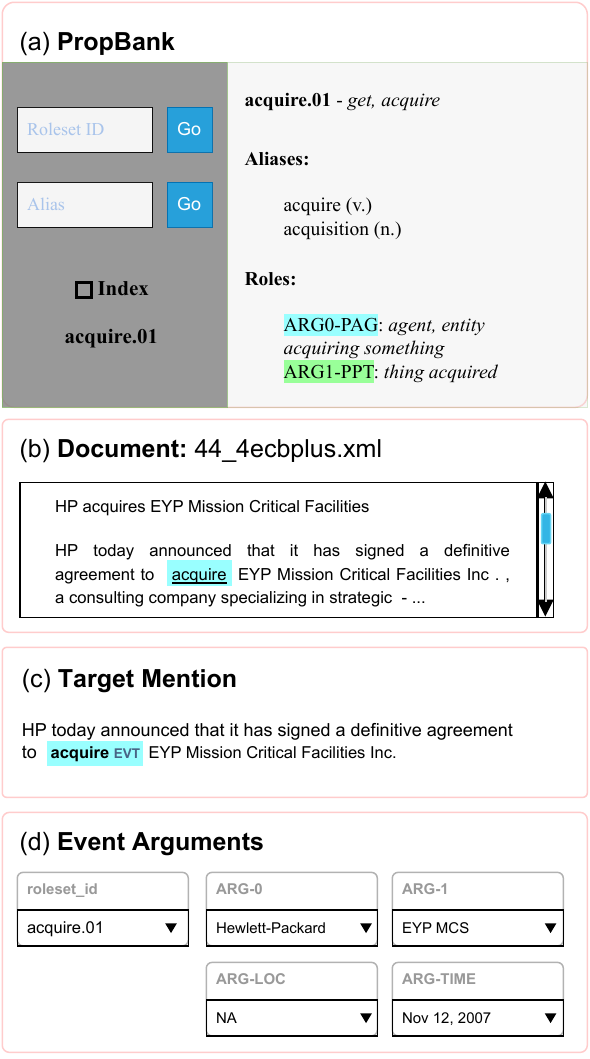}
    \caption{The Annotation Interface Using prodi.gy Annotation Tool
    }
    \label{fig:anno-ui}
\end{figure}

\subsection{Annotation Interface}

The annotation interface, as depicted in Figure \ref{fig:anno-ui}, comprises four distinct components: (a) the integrated PropBank website, (b) the document view, (c) the sentence view, and (d) the event argument forms. This interface is hosted on a server using Prodigy, with links distributed to individual annotators.

\vspace{1.5mm}
\noindent \textbf{PropBank Website}:
We adapt the publicly available PropBank website builder\footnote{https://github.com/propbank/propbank-frames} to ensure compatibility within an embedded environment. This interactive website hosts an indexed list of roleset definitions that annotators refer to.

\vspace{1 mm}
\noindent \textbf{Target Mention Document}:
The document containing the current mention is fully displayed in a scrollable view with the event trigger highlighted upon interface loading, facilitating easy access to additional context for annotators.

\vspace{1 mm}
\noindent \textbf{Target Mention Sentence}: This section displays the sentence encompassing the mention, with the event trigger highlighted in Prodigy's named entity recognition (NER) style. Typically, a sentence alone is sufficient to identify the arguments, and therefore, it is in the field of focus first.

\begin{figure}[t]
    \centering
    \includegraphics[scale=0.53]{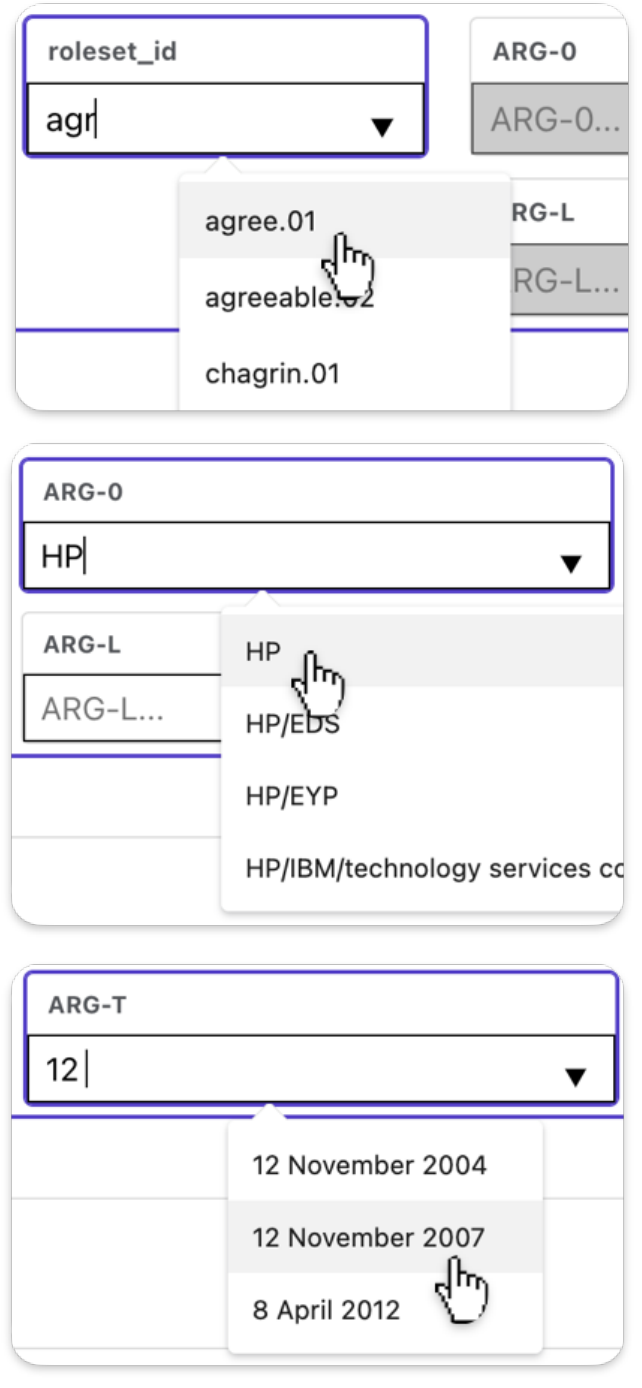}
    \caption{Screenshots}
    \label{fig:screenshots}
\end{figure}

\vspace{1 mm}
\noindent \textbf{Event Arguments Forms}: The event argument forms are located in this section, enabling annotators to manually input corpus-level arguments for the events. Each form is equipped with a dropdown list containing previously annotated arguments, facilitating the annotation process. Figure \ref{fig:screenshots} shows the different kinds of arguments stored in each of the argument forms. The \texttt{roleset\_id}~form stores all the rolesets in PropBank, \texttt{ARG-0}~and \texttt{ARG-1}~the identified agents and patients up til then, \texttt{ARG-LOC}~the locations, and \texttt{ARG-Time}~the dates.

\subsection{Model-in-the-loop}
Incorporating a model-in-the-loop approach, our annotation framework utilizes a straightforward Word2Vec classifier implemented using spaCy. This classifier ranks sentences containing previously seen arguments in relation to the target sentence. The dynamic ranking of these sentences is reflected in the dropdown list, with the highest-ranked sentence positioned at the top. The annotator is presented with the option to either accept or reject the top-ranked arguments.

\noindent \textbf{Argument Ranking and Selection:}
Upon loading the annotation interface, the system ranks the arguments from previously annotated sentences alongside the target sentence. The highest-ranked argument is selected by default and presented as the initial choice to the annotator. This ranking is based on the similarity or relevance of the sentences as determined by the Word2Vec classifier.

\noindent \textbf{Acceptance and Integration:}
Should the annotator choose to accept the top-ranked sentence, it is seamlessly integrated into the set of previous arguments. This integration enhances the corpus-level annotation by incorporating contextually relevant information from the selected sentence.

\noindent \textbf{Rejection and New Argument Creation:}
In the event of rejection, the system generates new arguments, leveraging the embedding of the rejected sentence. This adaptive mechanism ensures that even when an annotator rejects the top-ranked sentence, valuable information is not lost. Instead, it is used to generate potentially relevant arguments for further annotation.

\noindent \textbf{GPT-in-the-loop: }
Finally, yet importantly, we employ a GPT-based methodology to streamline the extraction of cross-document arguments through a two-step Retrieval Augmented Generation process. A comprehensive breakdown of our prompt engineering techniques is provided in Appendix \ref{sec:app-prompt}. The primary objective of this approach is to establish cross-document entity coreference.

Because of budget constraints, we have limited the execution of this experiment to a subset of the Dev dataset (Dev-small), encompassing a total of 120 mentions. Corpus statistics and annotation analysis are detailed in  Appendix \ref{sec:app-data}.
\section{Analysis}
\subsection{Model-in-the-loop}
We collect X-AMR annotations on the ECB+ dataset, as detailed in Appendix \ref{sec:app-data} (refer to the appendix for specific numerical data and human annotation analysis). During the annotation process, we collect human annotations along with predicted rolesets and arguments generated by our model. We assess the model's performance by comparing its predictions to human annotations. We carefully recorded the instances in which annotators made modifications to the predicted text provided by the model. We count the acceptance ratio of the predictions, which not only signifies the model's effectiveness but also represents the amount of effort saved by annotators.

\begin{figure}[t]
    \centering
    \includegraphics[scale=0.485]{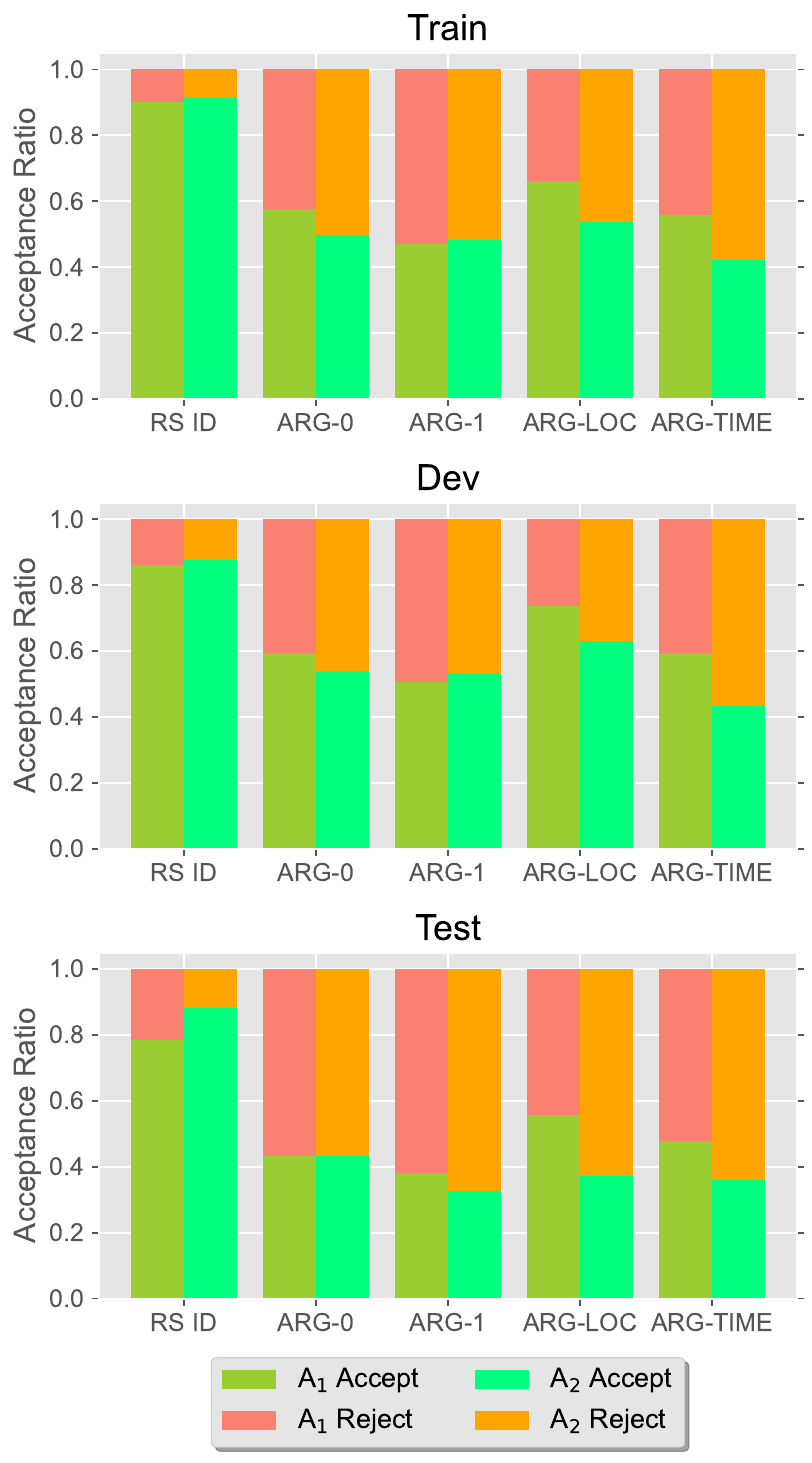}
    \caption{Roleset and ARG Analysis for A$_{1}$ and A$_{2}$: ``A$_{x}$ Accept'' represents the acceptance rate of the model suggestions according to Annotator $x$; ``A$_{x}$ Reject'' represents the rejection rate of the model suggestions according to Annotator $x$;}
    \label{fig:arg-ratios}
\end{figure}

Our analysis on the train, dev, and test sets of ECB+, as illustrated in Figure \ref{fig:arg-ratios}, reveals several noteworthy observations: the correct roleset ID prediction consistently exceeded 80\% for both annotators, denoted as A1 and A2. A1 appeared to be more inclined to accept the model's argument predictions compared to A2. This experiment serves as a foundation for future research, and one potential avenue is to incorporate these findings into downstream tasks, such as Event Coreference Resolution, to evaluate the quality of annotations and explore further implications of using model-in-the-loop for X-AMR annotations.
\begin{figure}[b!]
    \centering
    \includegraphics[scale=0.49]{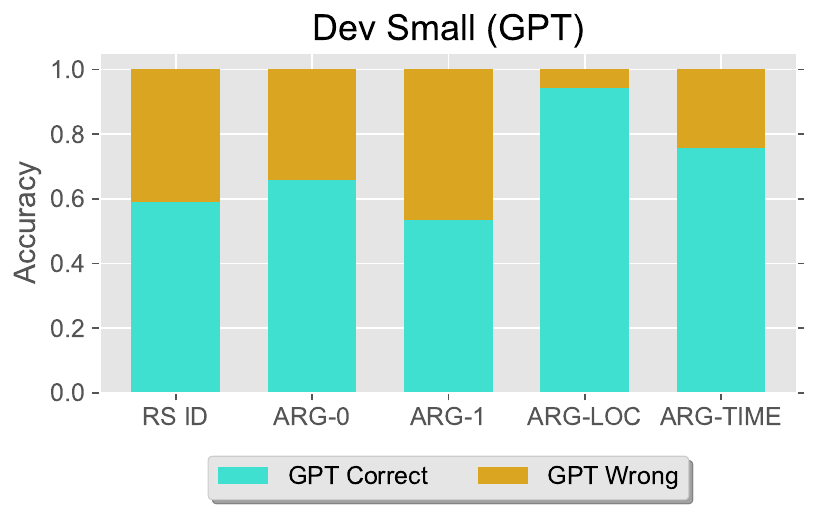}
    \caption{Accuracy of GPT Predictions of Roleset and ARG based on the gold standard annotation (adjudicated); }
    \label{fig:gpt-ratios}
\end{figure}
\subsection{GPT-in-the-loop}
In our GPT experiment on Dev-small, we had an adjudicator review 120 mentions and note when they had to adjust GPT's predictions. The outcomes of this evaluation are visually represented in Figure \ref{fig:gpt-ratios}, which illustrates the ratio of mentions requiring modification. The main takeaway here is that GPT performed well in generating Location and Time arguments but struggled with predicting roleset IDs and ARG-0, ARG-1 arguments. We believe that integrating the model-in-the-loop approach could help improve performance compared to just using GPT.

\section{Future Work}
The next steps include leveraging the X-AMR structures in creating efficient methods for neuro-symbolic event coreference resolution (ECR). For example, the X-AMR annotations could help in filtering the most pertinent event pairs that can be used with more resource intensive methods for estimating coreference \cite{ahmed-etal-2023-2}. Another important direction is in the estimation of the quality and cost savings of our methodology in doing ECR annotations. Quality measured by the number of ECR links that can be found with the least amount of pairwise event mention comparisons \cite{ahmed-etal-2023-good}.
\section{Conclusion}
In this paper, we have introduced a novel approach for cross-document, corpus-level semantic event extraction utilizing the X-AMR framework. To facilitate this process, we have developed a model-in-the-loop annotation tool tailored for X-AMR annotation, seamlessly integrated with Prodigy. This tool has been employed to curate X-AMR annotations by enriching an existing event coreference dataset, with contributions from two annotators. To evaluate the effectiveness of our approach, we have introduced a comprehensive assessment of the predictions, incorporating both the model's output and the assistance of GPT.


\section*{Limitations}
This work has several limitations. Firstly, the annotation tool used is a one-time paid software, which may restrict its accessibility to some researchers, although we have made the annotation recipe freely available. Secondly, the study relies on gold mentions rather than predicted ones, suggesting a need for future research to incorporate an additional annotation process to identify event triggers. Lastly, the non-reproducibility of GPT is acknowledged, and it may have been pre-trained on the corpus. However, we provide GPT-generated outputs and use them primarily for information generation rather than prediction, especially in event description generation. Future work may focus on distilling information into smaller, reproducible models to address these limitations and enhance the robustness of our approach.

\section*{Ethics Statement}
Recognizing the rigor and tediousness of the annotation process, our research ensured that all annotators were fairly compensated, given reasonable work hours, and provided with regular breaks to maintain consistency and quality. Comprehensive training and clear guidelines were offered, and a robust communication channel was established to address concerns, ambiguities, and to encourage feedback. Our team made efforts to involve a diverse group of annotators to minimize biases.

To alleviate the monotonous nature of the task, we employed user-friendly tools, rotated tasks, and supported peer discussions. We also acknowledged the crucial role of annotators in our research, ensuring their contributions were recognized and valued. Post-task, a summary of our findings was shared with the annotators, incorporating their feedback into the final manuscript, underlining our commitment to an inclusive and ethical research approach.

By adhering to the EACL guidelines, we aim to emphasize the ethical considerations surrounding the involvement of annotators in research projects. We believe that a humane, respectful, and inclusive approach to data annotation not only results in superior-quality datasets but also upholds the dignity and rights of all involved.

\section*{Acknowledgements}
The authors would like to thank the reviewers of EACL 2024 System Demonstrations who helped improve this paper. Part of this work was done during an internship of one of the authors at ExplosionAI GmbH. We would also like to thank Ákos Kádár, Matthew Hannibal, and the BoulderNLP group for their valuable comments on this paper. We gratefully acknowledge the support of  DARPA FA8750-18-2-0016-AIDA – RAMFIS: Representations of vectors and Abstract Meanings for Information Synthesis and a sub-award from RPI on DARPA KAIROS Program No. FA8750-19-2-1004.  Any opinions, findings, conclusions, or recommendations expressed in this material are those of the authors and do not necessarily reflect the views of DARPA or the U.S. government.
\bibliography{anthology,custom}
\bibliographystyle{acl_natbib}

\appendix

\section{Dataset Details}
The ECB+ corpus  is
a popular English corpus used to train and evaluate
systems for event coreference resolution. It extends
the Event Coref Bank corpus (ECB; \citet{bejan-harabagiu-2010-unsupervised}), with annotations from around
500 additional documents. The corpus includes
annotations of text spans that represent events, as
well as information about how those events are
related through coreference. We divide the documents from topics 1 to 35 into the training and
validation sets2
, and those from 36 to 45 into the
test set, following the approach of \citet{cybulska-vossen-2014-using}.
\label{sec:app-data}
\subsection{Annotation Analysis}
We have currently annotated all the mentions in the corpus with their Roleset IDs and 5,287 out of the 6,833 with X-AMR. In the three splits, only the Dev set has been fully annotated. We calculate the inter-annotator agreement (IAA) on the common Roleset predictions. The IAA is highest for the Dev set at 0.91, as depicted in Table \ref{tab:ecb}.

\renewcommand{\arraystretch}{1.4}
\begin{table}[htb]
\centering
\small
    \begin{tabular}{@{}ccccc|cc@{}}
     \toprule
    \multicolumn{2}{c}{~} & \textbf{Train} & \textbf{Dev} & \multicolumn{1}{c}{\textbf{Test}} & \makecell{\textbf{Dev}\\\textbf{small}}\\

    \cline{1-7}
	Documents && 594 & 196 & 206   & 91&\\
	Mentions  && 3808 &  1245 & 1780& 120 &\\
        \cline{1-7}
        \makecell{ \\[-3mm] Roleset ID \\ Agreement} && 0.84 & \textbf{0.91} & 0.80 & -- &\\
        \cline{1-7}
        w/ X-AMR && 3195$^{*}$ & 1245 & 847$^{*}$ & 120 &\\
        w/ Nested ARG-1 && 1081 & 325 & 220 & 24&\\
        w/ ARG-Loc && 2949 & 1243 & 707 & 120&\\
        w/ ARG-Time && 3192 & 1244 & 805 & 120&\\ \cline{1-7}
        \bottomrule
	\end{tabular}
  \caption[\ecb~Corpus Statistics]{
	ECB+ Corpus statistics for event mentions in \ecb~and the mentions annotated with X-AMR ($^*$Annotation in Progress). Inter-annotator agreement for the Roleset ID is highest for the Dev set. }
\label{tab:ecb}
\end{table}

\noindent \textbf{Arguments}:
Our analysis reveals a significant presence of mentions with nested ARG-1 annotations, as highlighted in Table \ref{tab:ecb} (w/ Nested ARG-1). This underscores the importance of capturing nested event relationships effectively. Additionally, our annotations for location and time modifiers successfully capture this information for nearly all mentions (w/ X-AMR), thanks to the assistance provided by drop-down options and the model-in-the-loop approach. These tools are particularly valuable in cases where date references are not explicitly mentioned in the document.
\section{Prompt Engineering}
\label{sec:app-prompt}
Our approach for X-AMR extraction with GPT involves a two-step process. In the initial step, we extract the Event Description along with the document-level arguments of the event by utilizing prompts such as \texttt{Instructions A}, \texttt{JSON Labels A}, and \texttt{Inputs A}. Following the generation of individual event descriptions through this step, we employ another prompt-based technique to generate corpus-level arguments.

In this secondary method, we introduce an additional instruction into \texttt{Instructions A}, forming \texttt{Instructions B}. This instruction directs GPT to identify the most informative Event Description that is coreferent with the current Event. Subsequently, we provide this identified Event Description (\texttt{JSON Labels B}) within the context and task GPT with generating missing information, such as date and location, pertaining to the target event. We provide the list of informative event descriptions in the topic of the target event in \texttt{Inputs B}.

The estimated cost of running this experiment is about \$15.

\newtcolorbox{mybox}[1][]{enhanced,colback=green!5!white,
colbacktitle=green!85!black!70!white,
colframe=green!75!black,fonttitle=\bfseries,
underlay={\begin{tcbclipinterior}
\end{tcbclipinterior}},
attach boxed title to top center={yshift=-2mm,xshift=-15mm},#1}

\newtcolorbox{mybox2}{
  enhanced,colback=green!85!black!70!white,        
  colframe=green!75!black,        
  coltext=fontcolor,       
}

\begin{mybox}[title=\tt Instructions A]
\small
\tt
\hspace{-1pt}You are a concise annotator that follows these instructions:
\vspace*{-1.5mm}
\begin{enumerate}[leftmargin=*]

\setlength{\itemsep}{2pt}
  \setlength{\parskip}{0.5pt}
    \item Identify the target event trigger lemma and its correct roleset sense in the given text.
    \item Annotate the document-level \argzero\space and \argone\space roles using the PropBank website for the roleset definitions.
    \item If the \argone\space role is an event, identify the head predicate and provide its roleset ID.
    \item Perform within-document and cross-document ana\-phora resolution of the \argzero\space and \argone\space using Wikipedia.
    \item Use external resources, such as Wikipedia, to annotate \argL\space and \argT .
\end{enumerate}
\end{mybox}

\begin{mybox}[title=\tt JSON Labels A]
\small
\tt
\hspace{-1pt}Here are the definitions of the keys in the JSON output:
\vspace*{-1.5mm}
\begin{itemize}[label={}, leftmargin=1pt]
\setlength{\itemsep}{1pt}
  \setlength{\parskip}{0.3pt}
    \item \textbf{Roleset ID}: The PropBank Roleset ID corresponding to the event trigger
    \item \textbf{ARG-0}: The text in the Document corresponding to the typical agent
    \item \textbf{ARG-0 Coreference}: The reference to the ARG-0 in Wikipedia in the format /wiki/Wikipedia\_ID
    \item $\vdots$
    \item \textbf{ARG-1 Roleset ID}: If the Event is Nested, provide the Roleset ID for the head event in ARG-1 clause
    \item \textbf{ARG-Location}: The reference to the event location in Wikipedia
    \item \textbf{ARG-Time}: The event time in the format of Month-Day-Year in your knowledge of the world or the document
    \item \textbf{Event Description}: In a single sentence, summarize the event capturing the Roleset\_ID and the names and wiki links of the Participants, Location and Time
\end{itemize}
\end{mybox}

\begin{mybox}[title=\tt Inputs A]
\small
\tt
\vspace{2mm}
\begin{itemize}[label={}, leftmargin=1pt]
\setlength{\itemsep}{1pt}
  \setlength{\parskip}{0.3pt}
    \item \textbf{Target Mention Document}: Entire document with the marked event trigger
    \item \textbf{Target Mention Sentence}: Sentence with the marked event trigger
\end{itemize}
\end{mybox}

\begin{mybox}[title=\tt Instructions B]
\tt
\vspace{2mm}
\begin{mybox2}
\normalsize
\textbf{Instructions A}
\end{mybox2}
\small
    6. Identify the most informative (having Wikipedia and complete dates) and best matching Event Description from the provided list of descriptions.

\end{mybox}

\begin{mybox}[title=\tt JSON Labels B]
\tt
\vspace{2mm}
\begin{mybox2}
\textbf{JSON Labels A}
\end{mybox2}
\small
\textbf{Most Informative Event Description}: Pick the most informative event description from the Event Description List. Choose by selecting the one that has complete date and Wikipedia links for the arguments and also is coreferent with the target Event. Hint: choose the one starts starts with "On DATE"

\end{mybox}

\begin{mybox}[title=\tt Inputs B]
\vspace{2mm}
\small
\tt
\begin{itemize}[label={}, leftmargin=1pt]
\setlength{\itemsep}{1pt}
  \setlength{\parskip}{0.3pt}
    \item \textbf{Event Description List}: Event descriptions of the three most informative and similar events in the corpus.
    \item \textbf{Target Event Description}: Event description of the target event
    \item \textbf{Target Mention Sentence}: Sentence with the marked event trigger
\end{itemize}
\end{mybox}

\end{document}